\documentclass[10pt,twocolumn,letterpaper]{article}

\usepackage[final]{cvpr}      % To produce the REVIEW version
\usepackage{bbding}
\usepackage{booktabs}
\usepackage{todonotes}
\usepackage{amsmath, amssymb}
\usepackage{stfloats}
\usepackage{float}
\usepackage{graphicx}
\definecolor{cvprblue}{rgb}{0.21,0.49,0.74}
\usepackage[pagebackref,breaklinks,colorlinks,citecolor=cvprblue]{hyperref}
\usepackage{multirow}

\def\paperID{2178} % *** Enter the Paper ID here
\def\confName{CVPR}
\def\confYear{2024}

\title{Uni3D-LLM: Unifying Point Cloud Perception, Generation and Editing with Large Language Models}

\author{
    Dingning Liu$^{1,2}$\\
    \tt \small{dutldn@mail.dlut.edu.cn}
    \and Xiaoshui Huang$^{1}$\footnotemark[1]\\
    \tt \small{huangxiaoshui@pjlab.org.cn}
    \and Yuenan Hou$^{1}$
    \and Zhihui Wang$^{2,}$\footnotemark[1]
    \and Zhenfei Yin$^{1}$
    \and Yongshun Gong$^{3}$
    \and Peng Gao$^{1}$
    \and Wanli Ouyang$^{1}$
    \and $^{1}${Shanghai Artificial Intelligence Laboratory}
    \and $^{2}${Dalian University of Technology}
    \and $^{3}${Shandong University}
}

\begin{document}
\maketitle
\begin{abstract}
\begin{figure*}[h]
  \centering
   \includegraphics[width=1\linewidth]{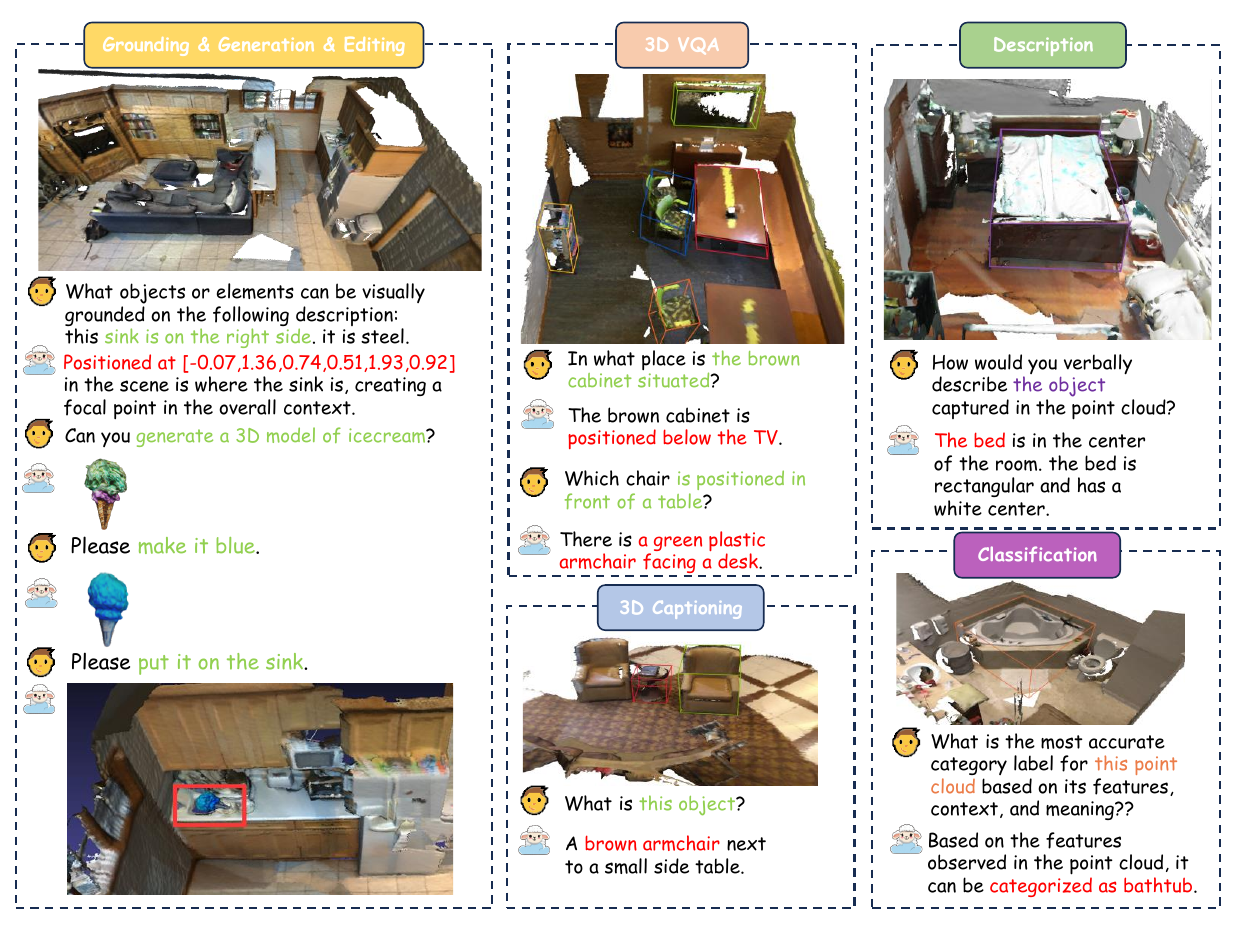}

   \caption{The example of Uni3D-LLM. Uni3D-LLM can complete various point cloud tasks in a unified framework, including 3D perception, point cloud generation and editing.}
   \label{fig:overview}
\end{figure*}

In this paper, we introduce \textbf{Uni3D-LLM}, a unified framework that leverages a Large Language Model (LLM) to integrate tasks of 3D perception, generation, and editing within point cloud scenes. This framework empowers users to effortlessly generate and modify objects at specified locations within a scene, guided by the versatility of natural language descriptions. \textbf{Uni3D-LLM} harnesses the expressive power of natural language to allow for precise command over the generation and editing of 3D objects, thereby significantly enhancing operational flexibility and controllability. By mapping point cloud into the unified representation space, Uni3D-LLM achieves cross-application functionality, enabling the seamless execution of a wide array of tasks, ranging from the accurate instantiation of 3D objects to the diverse requirements of interactive design. Through a comprehensive suite of rigorous experiments, the efficacy of Uni3D-LLM in the comprehension, generation, and editing of point cloud has been validated. Additionally, we have assessed the impact of integrating a point cloud perception module on the generation and editing processes, confirming the substantial potential of our approach for practical applications.
\end{abstract}
    
\section{Introduction}
\label{sec:intro}

In recent years, multimodal large language models (MLLMs) have made significant strides in the fields of natural language processing and computer vision. The powerful language capabilities of MLLMs enable them to handle various textual and visual tasks. Extensive research work~\cite{alayrac2022flamingo,li2023blip,liu2023visual,ye2023mplug,zhu2023minigpt,huang2023voxposer,zhang2022pointclip,huang2022frozen} has demonstrated the ability of MLLMs to integrate natural language with multiple modalities, including images and point clouds. The advancements in this integration technique have brought tremendous potential to applications, such as precise spatial analysis, augmented interactive experiences in augmented reality, and automated design.

Currently, there have been some approaches to explore the integration of additional functionalities into 3D scenes using MLLMs. These works can be classified into three main categories, \ie, direct embedding \cite{xu2023pointllm}, 2D-to-3D mapping \cite{hong20233d} and pre-alignment \cite{guo2023point}. As to direct embedding, PointLLM \cite{xu2023pointllm} utilizes pre-trained encoders to directly embed the features of point clouds into the textual space, which facilitates detailed object descriptions and classification tasks. As for the second class of methods, 3D-LLM \cite{hong20233d} reconstructs 3D features by generating rendered images from different perspectives. These reconstructed features are aligned with textual data and utilized for tasks such as visual question answering (VQA), planning, and dialogue in indoor scene settings. The third line of methods, as demonstrated in the work of Point-Bind~\cite{guo2023point}, typically align point clouds with other modalities such as images in multi-modal datasets. Subsequently, task-specific heads are employed for textual interaction and object generation. These exploratory methods provide valuable insights and practical approaches for integrating additional functionalities into 3D scenes.

Previous approaches, however, still exhibit certain limitations. Firstly, while aligning point clouds with text directly may seem straightforward, accurately recognizing and comprehending the spatial information of the point cloud poses a challenge for LLM. This limitation restricts its capacity to generate precise scene-level interpretations and descriptions. Secondly, although the conversion from 2D to 3D can reconstruct 3D features, it may encounter orthographic occlusion issues when dealing with images captured from different viewpoints. Consequently, this can lead to incomplete 3D scene reconstruction and underutilization of the original point cloud features. Thirdly, although Point-Bind \cite{guo2023point} aligns point clouds with other modalities, it is limited to achieving object-level alignment. Aligning point clouds at a scene level with other modalities presents a significant challenge. Moreover, Point-Bind treats LLM and the generation model as separate downstream tasks, overlooking the crucial influence of deep semantic understanding inherent in LLM on the controllability of the generative procedure.

%Although there are already various methods mentioned above utilizing large language models (LLMs) to process various point cloud tasks, none of them have considered the necessity to integrate 3D perception, generation, and editing. The existing methods are confronted with great difficulties in accomplishing this series of tasks, such as scattered processing workflows, low efficiency, and high costs. The aforementioned shortcomings motivate us to recognize the importance of unifying the application of LLMs in 3D perception, generation and editing. The integration not only can address the shortcomings of traditional 3D perception and generation tasks but also significantly enhances collaborative work and overall efficiency. With just a single training, it becomes possible to facilitate mutual enhancement across various scenarios.  In text-to-3D generation, LLMs can grasp richer semantic information to guide the generation process, and in the meanwhile, their advancements in perception also provide a solid foundation for more accurate generation and editing. Furthermore, this unified framework also promotes interaction between different tasks, making iteration and refinement in complex projects more efficient.
Despite the progress achieved by several preliminary attempts utilizing large language models (LLMs) for point cloud tasks , none of them have considered integrating 3D perception, generation and editing into a unified framework. 
The existing methods are confronted with significant challenges in accomplishing this series of tasks, including fragmented processing workflows, low efficiency, and not utilizing the rich semantic knowledge of LLM to generate more freely. But these challenges involve not only enhancing the perception capabilities of 3D scenes but also bridging the gaps between different modalities and tasks, and effectively integrating the rich linguistic information of Large Language Models (LLMs) into generation models. 
These shortcomings highlight the importance of unifying the application of LLMs in 3D perception, generation, and editing. 
A unified framework not only addresses the limitations of traditional 3D perception and generation tasks, but also greatly enhances collaborative work and overall efficiency. With a single training recipe, mutual enhancement across various scenarios becomes possible.
In the text-to-3D generation, LLMs can leverage rich semantic information to guide the generation process. Additionally, their advancements in perception provide a solid foundation for more accurate generation and editing. Moreover, this unified framework promotes interaction between different tasks, enabling more efficient iteration and refinement in complex projects.

%Unifying the processes of 3D perception, generation, and editing poses a significant challenge in our research.  This challenge involves not only enhancing the perception capabilities of 3D scenes but also bridging the gaps between different modalities and tasks, and effectively integrating the rich linguistic information of Large Language Models (LLMs) into generation models. 
%In this paper, we propose a novel unified framework, Uni-3DLLM. Uni3D-LLM focuses not only on how LLMs can more deeply understand and process the details of 3D environments but also emphasizes using the linguistic capabilities of LLMs to enrich and guide the generation of 3D content. Additionally, we introduce an LLM-guided method for 3D generation to editing. By fusing scene point clouds and image information to align both modalities with text, we complete the perception task of point clouds. We designed an information mapping module to transfer the rich semantic features of LLMs to the generation model, achieving accurate generation. Subsequently, using modified rendering images from different angles, we update the original 3D model step by step to obtain a new, edited version. Our model is not only versatile in various scenarios but also enables efficient mutual enhancement during generation and editing processes.
In this paper, we introduce a novel unified framework called Uni-3DLLM, which aims to enhance the understanding and processing of 3D environments through the utilization of large language models (LLMs). Uni-3DLLM not only focuses on enabling LLMs to delve deeper into the details of 3D environments but also leverages their linguistic capabilities to guide the generation of 3D content.
We propose an LLM-guided method for 3D generation and editing within this framework. By integrating scene point clouds and image information and aligning them with text, we accomplish the perception task of point clouds. To facilitate accurate generation, we design an information mapping module that transfers the rich semantic features of LLMs to the generation model. Subsequently, by iteratively updating the original 3D model using modified rendering images from various angles, we obtain a new, edited version.
Our proposed model exhibits versatility across different scenarios and enables efficient mutual enhancement during both the generation and editing processes.

%Specifically, Uni3D-LLM employs 3D detection algorithm to process scene point clouds, transforming them into features of multiple Regions of Interest (ROI). These ROI features are then concatenated and integrated into a complete scene point cloud input. Simultaneously, the top-to-down view of the related point cloud scene, serving as an auxiliary input, is also mapped into the textual space for perception tasks. In the generation task, extra learnable generation tokens are added at the end of the generated descriptions. These tokens are then transformed through our mapping block into understandable signals by the Generator. The world knowledge contained in LLM endows our method with the ability to generate and edit objects, even if user descriptions are rough or vague. During the training phase, we adopted a two-stage separated approach. First, we integrate the generation mapping module into LLM and then freeze LLM. Then, we train the LLM to access the information features of point clouds through Parameter Efficient Fine Tuning (PEFT). This strategy effectively prevents catastrophic forgetting, ensuring robust and coherent learning throughout the entire training process.
% 是怎么做的，来实现了unified framework，做了什么特殊的操作能够克服之前的不足（具体的）
Specifically, within the Uni3D-LLM framework, to obtain better scene-level point cloud feature, we replace the original point cloud with a combination and integration of features from each object in the scene. In addition, We utilize various powerful image encoders to extract the top-down view features of the point cloud scene and embedd them into the textual space. 
In the generation task, in order to make the generation model understand the semantics of LLM, we introduce extra learnable generation tokens at the end of the language descriptions. These tokens are transformed through our mapping block, enabling the Generator to generate understandable signals. The world knowledge contained within the LLM empowers our method to generate and edit objects, even when user descriptions are rough or vague. And during the training phase, to connect perception, generation and editing, we adopt a two-stage approach. Firstly, we integrate the generation mapping module into the LLM and freeze the LLM. Then, we train the LLM to access the information features of point clouds using Parameter Efficient Fine Tuning (PEFT). This strategy effectively prevents catastrophic forgetting, ensuring robust and coherent learning throughout the entire training process.

The contributions of Uni3D-LLM are summarized as follows:\\
\begin{itemize}
    \item \textbf{A Unified framework to Process Multiple 3D Tasks with LLM.} We make the first attempt to employ LLM to unify a wide array of 3D tasks, including 3D object generation, editing, 3D perception, 3D visual grounding to solve the disconnect between user intent, conveyed through language, and the execution of 3D tasks, providing a more natural and fluid interaction paradigm.
    \item \textbf{Multimodal Signal Alignment.} We pioneered the use of point cloud and additional image assisted by carefully designed, modality-specific projectors, to map heterogeneous text, image, and point cloud signals into a common token space. The extracted multimodal tokens are fed into the LLM to generate rich semantic features, which are further sent to task-specific architectures to produce the desired outputs.
    \item \textbf{Multi-Task Synergy.} We conducted extensive experiments to verify the synergistic effects of unifying various 3D tasks and to pave the way for building 3D foundation models.
\end{itemize}

%-------------------------------------------------------------------------%

\section{Related Work}
\label{sec:related}
\subsection{Multi-modal Large Language Model}
As the impact and accessibility of Large Language Models (LLMs) continue to grow, there is a growing body of research dedicated to extending these pretrained LLMs to handle multimodal comprehension tasks. Some studies have explored training models from scratch using a large amount of image-text pairs~\cite{openai2023gpt4,alayrac2022flamingo,li2023blip,li2023otter} 
and applied them to downstream tasks such as visual question answering(VQA), captioning, and coarse/fine-grained understanding, followed by fine-tuning. Other researchers have connected pretrained visual models with pretrained LLMs, incorporating additional mapping modules like QFormers~\cite{li2023blip}. This approach leverages the perceptual abilities of pretrained visual models and the reasoning and generalization capabilities of LLMs.
In our work, since many studies have demonstrated the strong perceptual capabilities of LLMs in images~\cite{liu2023visual,zhang2023llama,gao2023llama}, we utilize two encoder(image and point cloud) to align with texts to assist in acquiring point cloud information and enhance the spatial understanding of LLMs for point clouds.

\subsection{3D Object Generation}
3D generation is a task aimed at creating realistic and diverse 3D models from different inputs (such as text, images, sketches, or point clouds). This task is challenging and requires a deep understanding of the shape, structure, texture, and semantics of 3D objects. The main methods currently include parametric methods and non-parametric methods. Parametric methods use predefined templates or primitives to represent 3D shapes, such as voxels, meshes, point clouds, or implicit functions~\cite{smirnov2020deep,palafox2021npms}. These methods can generate high-resolution and high-fidelity smooth continuous 3D models. However, these methods also have some limitations, such as high computational cost, fixed topology, or difficulty in handling complex geometry. Non-parametric methods use generative models to learn the distribution of 3D shapes from data, such as Generative Adversarial Networks (GANs) \cite{creswell2018generative}, Variational Autoencoders (VAEs) \cite{kingma2013auto}, Normalizing Flows \cite{yang2019pointflow}. However, these methods also face some challenges, such as mode collapse, decoupling, or evaluation issues or . Recently, with the rise and rapid development of diffusion models in the 2D field, more and more 3D generation research has begun to adopt diffusion models~\cite{radford2021learning, lin2023magic3d,nichol2022point,liu2023zero,liu2023syncdreamer,shi2023mvdream}. But the mainstream issue now is that the generation time is not high enough or the generation quality is not good enough. Liu et al. use 3D Gaussian~\cite{kerbl20233d} to reconstruct the entire 3D scene rapidly. In our research, we use dreamgaussian~\cite{tang2023dreamgaussian}, a model that utilizes 3D Gaussian for rapid 3D object modeling, as decoder to complete the generative network. The original dreamgaussian model uses CLIP \cite{radford2021learning} to embed text, and CLIP is trained on billions of text-image pairs. Therefore, when performing text-to-3D conversion, users can usually only generate the 3D objects they expect by providing relatively short text prompts, and cannot achieve more natural and descriptive words.

\subsection{3D Editing}
3D Shape Editing is also a challenging task that requires a deep understanding of shapes. Traditional methods use explicit deformations, while recent years have seen the widespread adoption of CLIP, accelerating attempts to build language-guided editing systems for both images and 3D shapes. As CLIP is trained with pairs of images and texts, most recent efforts have focused on 2D image editing \citep{bau2021paint,brooks2023instructpix2pix}.  For 3D, some works introduced a framework for synthesizing 3D shapes and scenes from texts \citep{chen2019text2shape}.  However, these methods mainly focus on generating 3D shapes rather than editing, which requires language-shape alignment to resolve given edit descriptions. To achieve more intuitive and fine-grained 3D shape editing, some works have explored language-based 3D shape manipulation \citep{michel2022text2mesh,achlioptas2022changeit3d}. These works use powerful vision-language models to generate mesh vertex deformations and colors or build a shape auto-encoder in a latent space and a neural listener to edit shapes according to text instructions. However, these works still have some limitations, such as relying on predefined parts or object localization. In contrast, we propose a MLLM-based 3D editing framework that uses instruct tuning to perform 3D object shape editing with more advanced natural language.
%3D perception, cls, seg, det, grounding Uni series, uni-perceiver, uni-IO, PointLLM\&PointBind
\section{Method}
\label{sec:method}
\begin{figure*}[h]
  \centering 
   \includegraphics[width=1\linewidth]{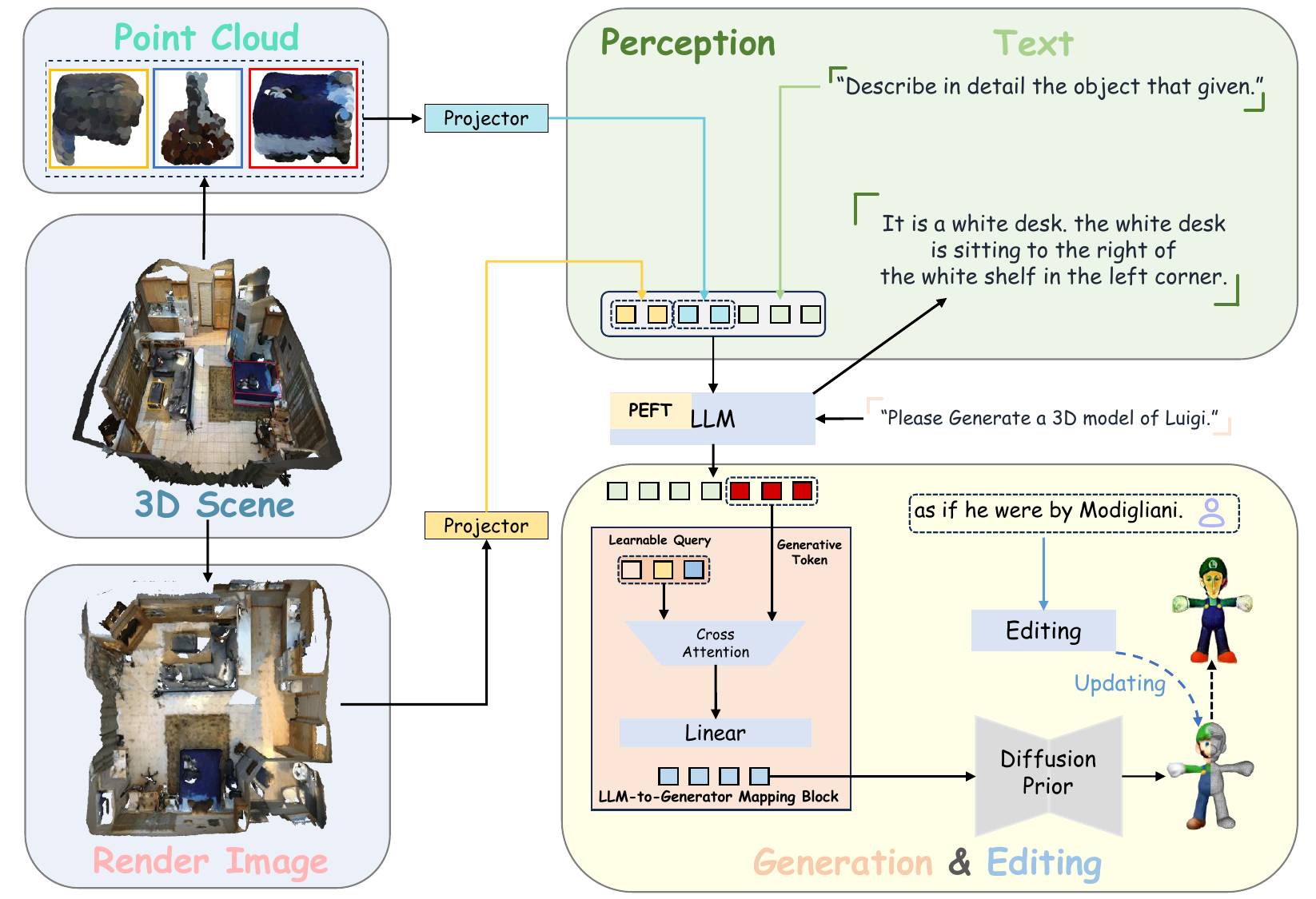}
   \caption{\textbf{The framework overview.} We first decompose the point cloud into sub point clouds using 3D detection algorithms and obtain its top-down view rendered image features. Both features added to complete various 3D perception tasks. Our instructions are passed to the generation and editing module through the mapping block.}
   \label{fig:overview}
\end{figure*}
In this part, we will present the model design of Uni3D-LLM
and the detail of training strategy.
\subsection{Model Design}
The overall framework of Uni3D-LLM is presented in \cref{fig:overview}. In the this section, we will describe the details of the multimodal input alignment layer, the LLM-to-Generator mapping block and the generation-editing module.
\subsubsection{Multi-modality alignment.} 
\begin{figure*}[h]
  \centering
   \includegraphics[width=0.8\linewidth]{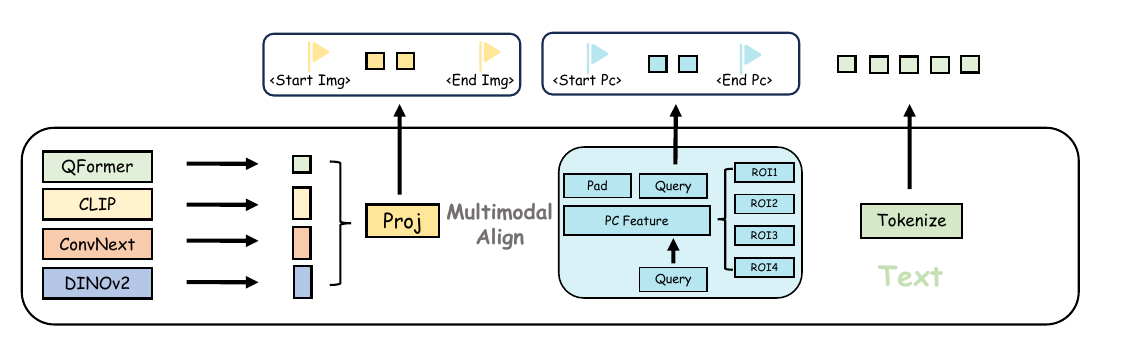}
   \caption{\textbf{Multimodal alignment method.} In which the images are re-connected together in the channel dimension after 3 encoder passes, in addition, the image is passing extra QFormer as a global feature concat with the other feature. the ROI is then recoded by Pointbert, with additional padding after cross attention in Query.}
   \label{fig:multi}
\end{figure*}

The multimodal alignment method is shown in the \cref{fig:multi}. Hong et al.~\cite{hong20233d} have established that the integration of 2D image data in the reconstruction of 3D information can be instrumental in augmenting the precision of point cloud recognition tasks. Consequently, in our approach, we have implemented a methodology that aligns point clouds with corresponding images data. To facilitate this cross-modal representation alignment, we have adopted modality-specific projector-based structures.
\begin{itemize}
\item \textbf{For the alignment of point cloud modality,} inspired by Octavius \cite{chen2023octavius}, we adopt a two-step approach to align the point cloud with the space of LLM. Firstly, we follow the detection method of rukhovich et al.~\cite{rukhovich2022fcaf3d} to extract the objects from the scene point cloud. Then, we employ a pretrained Point-Bert model \cite{yu2022point} as the encoder for extracting point cloud features. Furthermore, the cognitive module LLaMA2 is utilized to facilitate the alignment process. It is important to note that for different tasks, there are distinct approaches. For object-level tasks, the point cloud data is typically mapped into the textual space through a mapping layer. This enables the fusion of visual and textual information for object-level understanding. However, for scene-level tasks, such as grounding, a different strategy is employed. In this case, for each individual scene point cloud, additional position encoding is introduced to preserve its original spatial information. Subsequently, these encoded point clouds are recombined to form a cohesive representation of the entire scene point cloud input. This methodology ensures that the model captures both the local details and the global context of the scene, facilitating accurate scene-level tasks.
 \item \textbf{For the alignment of image,} we directly incorporate the 2D representation extraction method from sphinx. We utilize multiple pre-trained encoders~\cite{li2023blip,radford2021learning,dosovitskiy2020image,liu2022convnet,oquab2023dinov2} to extract global and local features from the images. For a given scene point cloud, we consider the potential occlusion and limited visibility caused by different rendering pose. To address this, we employ a top-view representation as a representation of image modality. Similarly, we utilize two learnable special tokens to indicate the beginning and end of the inserted image. This allows us to effectively capture the essential visual information of the scene while mitigating the impact of occlusion and incomplete visibility.  
Once the features for both modalities are extracted, we concatenate align them at the beginning of the textual feature sequence. The image feature, serving as the global guiding feature, is placed at the front. special tokens indicating the start and end of the respective modalities). This concatenation allows for effective integration and alignment of the image and point cloud modalities within the overall textual feature representation. \end{itemize}
This alignment facilitates the utilization of the complementary insights offered by both images and text, thus enhancing the granularity and overall accuracy of recognition tasks within point cloud scenes.
\subsubsection{LLM-to-Generation mapping block.} To connect the language model output feature and the generation model, we establish a mapping block between them. 
During the training phase, when we input a generated text, we append 259 learnable generative tokens at the end, representing the desired image to be generated. In the final output feature, we extract the generative tokens and pass them through the mapping block to convert them into the corresponding generation features.
The language model, with its rich semantic understanding, serves as a powerful control mechanism for the generation process. It consists of a learnable query, transformer layers and MLPs as the projector to map the text features to the signal that can be known by the generation model. DreamGaussian~\cite{tang2023dreamgaussian} primarily utilizes Stable Diffusion~\cite{rombach2021highresolution} and SDS loss~\cite{poole2022dreamfusion} to guide the generation process of gaussian splatting\cite{kerbl20233d}. Our objective is to map our features as the text condition into Stable Diffusion. We aim to guide the diffusion process in a way that aligns with our desired outcomes. 
The overall process can be represented as follows:
\begin{equation}
 F_{gen} = \Theta_{\text{{map}}} \left( q, \Theta_{\text{{llm}}}(\langle\text{{Text}},  Token_{gen} \rangle) \right) \in \mathbb{R}^{L\times D}
  \label{eq:mappingblock}
\end{equation}
Where ${F_{gen}}$ represents the text feature fed into the generation model, ${\Theta}$ represents various network parameters, q represents the learnable parameters, L represents the acceptable vector length of the generation model, and D represents the dimension.\\
\subsubsection{Generation-to-editing module.}
\begin{figure*}[h]
  \centering
   \includegraphics[width=1\linewidth]{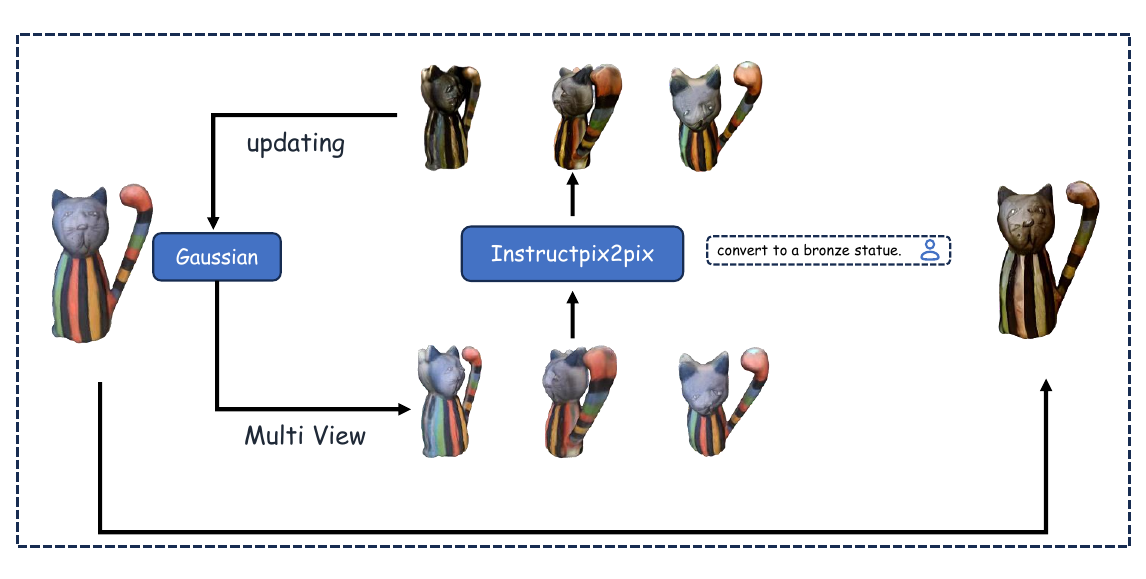}
   \caption{\textbf{The pipline of Editing method.} After we obtain a 3D model, we will also store the initialized 3D Gaussian. When the user is given the modification instruction, Generation-to-editing module will render multiple fixed perspective images and send them to Instructpix2pix and then update the 3D Gaussian gradually by updating the specified pose.}
   \label{fig:editing}
\end{figure*}

The overall editing process is shown in the \cref{fig:editing}. Once we obtain the generated 3D model, if we intend to modify the corresponding model, we can adopt a methodology similar to instruct nerf2nerf~\cite{haque2023instruct}. We utilize the generated Gaussian splatting data as the initial data and select several rendered images of the 3D model from different poses to make sure the consistency. We leverage instruct-pix2pix~\cite{brooks2023instructpix2pix} to generate the rendered images of the modified object. Subsequently, we gradually update the entire Gaussian splatting using the rendered images, ultimately completing the object editing process.
\subsection{Training Strategy}
\textbf{Stage I.} we first train our text-to-generation mapping block. In order to generate accurate images to guide 3D generation, the mapping feature ${F_{Gen}}$ plays a crucial role as a condition during the denoising process. The mapping feature is expected to capture the relevant text features that effectively guide the latent diffusion model (LDM) in generating the desired ground truth image. To achieve this, we leverage the LDM training loss as a guiding mechanism during training.
During training, the ground truth image is first encoded into a latent feature $f$ using a pretrained VAE. Subsequently, we add t steps noise ($\epsilon$) to the latent feature to obtain the noisy latent feature $f_t$. To calculate the conditional LDM loss, we utilize a pretrained U-Net model to predict the noise $\epsilon_{pred}$ added on the latent feature, which takes $f_t$ as input. The conditional LDM loss can be expressed as follows:
\begin{equation}
L_{LDM} = \mathbb{E}_{\epsilon \sim \mathcal{N}(0,1), t}(\epsilon, \epsilon_{pred}(f_t,t,F_{gen}))
\label{eq:LDMLoss}
\end{equation}
\textbf{Classifier-Free Guidance.}
To enhance the coherence between the text feature and the generator, we adopt the concept of Classifier-Free Guidance(CFG) for generation. We introduce a 10\% probability of replacing the mapping feature with zero features and the zero features will be as the negative prompt at the inference stage. \\
\textbf{Stage II.} Once the mapping layer has been trained, we proceed to train the perception module.Parameter Efficient Fine Tuning(PEFT)~\cite{houlsby2019parameter,hu2021lora} plays a crucial role in the training of LLMs and MLLMs.~\cite{chiang2023vicuna,zhu2023minigpt,yin2023lamm} In this study, we employ Lora on the LLM and during the training phase, the parameters of the whole large language model are frozen, and only the Lora layer is trained. By doing so, we can introduce new multimodal knowledge without losing the existing knowledge of the LLM.
\textbf{Implementation details.}
For the selection of large language models, we opted for sphinx\cite{lin2023sphinx}, an MLLM that incorporates the image modality on llama2\cite{touvron2023llama}. With this model, we can easily integrate the point cloud modality, thereby enhancing the performance of point cloud tasks by leveraging the joint information of point cloud and images. During training, all visual modality signals are embedded into the textual space as tokens of 259 length. The learning rate is set as $10^{-3}$ and the FuseAdam is chosen as the optimizer.
\section{Experients}
\label{sec:exp}
\begin{table*}[h]
\centering
\begin{tabular}{c|ccccc}
\hline \multirow{2}{*}{ Models } & VQA(Scanqa) &\multicolumn{2}{c}{ Classification(Acc@1)}  & Caption(Scan2Cap) & Grounding(Scannnet)\\ 
& BLEU-1& ShapeNet& Scannet& BLEU-1 & mAP@0.5\\
\hline 3D-LLM (Flamingo) & 30.30 & - & - & - & - \\
Octavius w/ MoE & 44.24 & 24.85 &  \textbf{48.80} & \textbf{35.94} & - \\
Ours(Only 3D) &  43.26 & 20.45 & 46.90 &33.60 & Failed \\
Ours(3D+img) &  \textbf{44.68} & \textbf{30.32} & 47.10 & 34.70 & \textbf{13.69} \\
\hline
\end{tabular}
  \caption{\textbf{Comparisons on 3D perception tasks.} we compared VQA, classification, caption task with 3D-LLM and Octavius. Among them, Scanqa, Scan2CAP, and Scannet are the results of fine-tuned.}
  \label{tab:perception}
\end{table*}

\begin{table*}[h]
\centering
\begin{tabular}{cccccccc}
\hline & \multirow{2}{*}{FID $\downarrow$}& \multicolumn{2}{c}{ClIP Score} &CLIP & \multicolumn{2}{c}{ R-precision } & \multirow{2}{*}{speed}\\
 & & prompt & descript & R@1 & R@5 & R@10 &\\
\hline Shape-E(NeRF) & 39.4 & 78.2 & 70.4 & 17.4 & 35.8 & 44.2& \~{2min} \\
Shape-E(STF) & 34.7 & 78.8 & 71.1 & 19.6 & 39.4 &47.5& \~{2min} \\
Uni3D-LLM & \textbf{33.4} & \textbf{79.7} & \textbf{71.3} & \textbf{22.3} & \textbf{42.6} &\textbf{52.4}& \~{3min} \\
\hline
\end{tabular}
  \caption{\textbf{Description-to-3D on Cap3descript.} Uni3D-LLM compared with \\ Shap-e after fine-tuning on Cap3d and Cap3descript.}
  \label{tab:gen}
\end{table*}
% ablation study
\subsection{Experimental Setup} To explore the effectiveness of our framework in multimodal learning, we fine-tune Uni3D-LLM in two modality setups: i.) only point cloud modality and ii.) both image and point cloud modalities. We then evaluate the zero-shot and fine-tuned performance using these two fine-tuned models on various 3D downstream tasks.\\
\textbf{Datasets.} For the training of perception, we utilize an instruction dataset called "Scan2Inst"~\cite{chen2023octavius} whitch generated from ScanNet~\cite{dai2017scannet} which consists of tasks such as description and classification. In addition, this dataset is also including Scanqa(VQA)\cite{azuma2022scanqa}, and Scan2Cap(Cap)\cite{chen2021scan2cap} for different task training data.
Moreover, we have also integrated additional grounding data into our training dataset. For each scene, we captured 1,513 top-down view rendered images of the scenes, which are used as supplementary information across all tasks. \\
For the training of our LLM-to-generator mapping block, we initially trained using 2D image-text pairs from the MS-COCO\cite{xu2018attngan} and LN-COCO\cite{yu2022scaling} datasets. Considering the target for 3D object generation, we also integrated pre-training with the Cap3d\cite{luo2023scalable}, a subset of 3D object-caption data extracted from Objaverse. This dataset includes 650k point cloud-different view render images-relevant description pairs. Additionally, to enable users to input descriptions as naturally as they would, we created a dataset called "Cap3descript" based on Cap3d. Since GPT cannot process point cloud data and the GPT4-V\cite{openai2023gpt4} API is not available, we used the open-source model PointBind-LLM\cite{guo2023point} to generate a dataset with 10,000 detailed descriptions based on Cap3D. For each point cloud data, we generated a set of descriptions for eight view images and the original point cloud. Subsequently, we used GPT4 to integrate these nine descriptions into one paragraph, serving as the complete description of the object.\\
\textbf{Implementation details.} we employ LoRA and task-specific learning when training the perception part. The rank of each LoRA is set to 32. We utilize the FusedAdam, an Adam optimizer\cite{adam2014method}, with a total batch size of 16 and the learning rate is $1 \times 10^{-4}$ for 2 epochs. All experiments are performed using 8 NVIDIA A100 GPUs.\\
The images are resized to 224$\times$224 and are processed by four different encoders, \ie, CLIP\cite{radford2021learning}, ConvNeXt\cite{woo2023convnext}, DINOv2\cite{oquab2023dinov2}, QFormer\cite{li2023blip}. For point cloud data, we extract regions of interest (RoI) using FCAF3D\cite{rukhovich2022fcaf3d} and sample 1024 points from each RoI. Each encoder is used for a pre-training weight. For each scene, we follow the settings that selecting N instances with a bounding box confidence higher than the threshold 0.3. In the multimodal fusion step, we employ 16 queries to obtain aligned 3D visual features. Additionally, we pad the output 3D visual features with masks to a size of 256, aligning with length of the image tokens.\\
\textbf{Quantitative Results.} The conclusions of all experiments are shown in the \cref{tab:perception}, \cref{tab:gen}.
For perception, we evaluate classification performance on ShapeNet~\cite{chang2015shapenet}, captioning performance on NR3D, vqa performance on ScanQA, and grounding on the test split of ScanNet.\\
For generation, We conducted two experiments: one is on the test set of Cap3d, and the other is on Cap3descript's test set which is curated test set of 100 samples extracted from the original test set, designed to assess the quality of generation under natural text descriptions. we evaluate the CLIP-Score on different fine-tuned 3D generation model to test the generation quality and time. Due to the time constraints for interactive use with LLM, excessively long generation times are not suitable. Therefore, we only fine-tuned the Shape-E and conducted a comparison. \\
For editing, it is challenging to have objective evaluation metrics to determine the quality. The assessment primarily relies on personal subjective judgment and feelings. Therefore, we do not evaluate the quality of our editing.\\
Based on \cref{tab:perception}, we can confirm that by introducing image auxiliary information, our model Uni3D-LLM is capable of performing grounding tasks. The primary reason for task failure when relying solely on point cloud information might be the sometimes inaccurate and incomplete capture of all objects in the Region of Interest (ROI), leading to fragmentation in the point cloud features. However, with the addition of complete image assistance, our model effectively handles questions and answers related to global scenes and specific objects.
We also observed that in the classification tasks of Scannet and the caption tasks of Scan2Cap, even with the addition of image assistance, there was no significant improvement in the model's performance. We analyze that the main reason might be that these tasks focus on individual properties, and images, as global auxiliary information, are insufficient for significantly aiding these individual-level Q\&A tasks.
Therefore, to further explore the role of image information in different tasks, we fine-tuned our model on the Cap3d and conducted zero-shot testing on ShapeNet. In these tests, we provided front-view images (view 5) of the objects from the dataset. The results showed that in the task of object-level point cloud classification, providing image information related to the object significantly enhances the overall classification results. This finding emphasizes the importance of combining object-level image information to enhance model performance in specific scenarios.
During the testing of generation performance, we experimented with both short prompts and longer, more natural text on \cref{tab:gen}. Considering that the CLIP Score is not highly accurate for long texts and images, we aligned the object generated by natural texts with their original caption. The results indicate that the overall generation still meets the expected outcomes.\\
\textbf{Ablation Study.}
\begin{table}
\centering
\begin{tabular}{ccc}
\hline Model & Add perception(Lora) & CLIP Score\\
\hline Ours & \XSolidBold & 79.5\\
Ours & \CheckmarkBold & \textbf{79.7}\\
\hline 
\end{tabular}
  \caption{\textbf{Ablation Study on Adding Perception.} In order to verify whether the multimodal sensing module interferes with the generation module, we use the test set and CLIP Score of the original Cap3d data set as the evaluation.}
  \label{tab:abl}
\end{table}
To investigate whether the introduction of the perception module would have any negative impact on the generation module, we conducted an ablation study. The experiment shown on \cref{tab:abl} was divided into two groups: the first group directly aligned the LLM using the LLM-to-Generator Mapping Block, while the second group performed the alignment after introducing the perception module, Lora. The results of the experiment show that the introduction of the perception module does not interfere with the generation results; in fact, it leads to a slight overall improvement in the performance metrics of the generation module.
% Quantitative results Visual comparisons Ablation study Elaborated analysis

\section{Conclusion and Limitations}
In this paper, we introduce Uni3D-LLM, the first attempt to integrate perception, generation, and editing for point clouds. By incorporating powerful image features as spatial assistance, we have overcome the perturbation issues in the original point cloud features that arise when solely inputting point cloud modality. Integrating multiple modalities as auxiliary information for point cloud tasks has proven beneficial for point cloud. However, further enhancing the positioning capability of point clouds remains a challenge for future research.Our generative-editing method also inherits many of the limitations of DreamGaussian and InstructPix2Pix, such as the inability to generate large-scale spatial scenes and perform more freeform directive editing operations. 
Our generative editing method also faces the same limitations as DreamGauss and InstructPix2Pix, such as the inability to generate large-scale spatial scenes and perform freeform editing. This also needs to be solved in the future.
\section{Details of Cap3descript}
\begin{figure*}[h]
  \centering
   \includegraphics[width=1\linewidth]{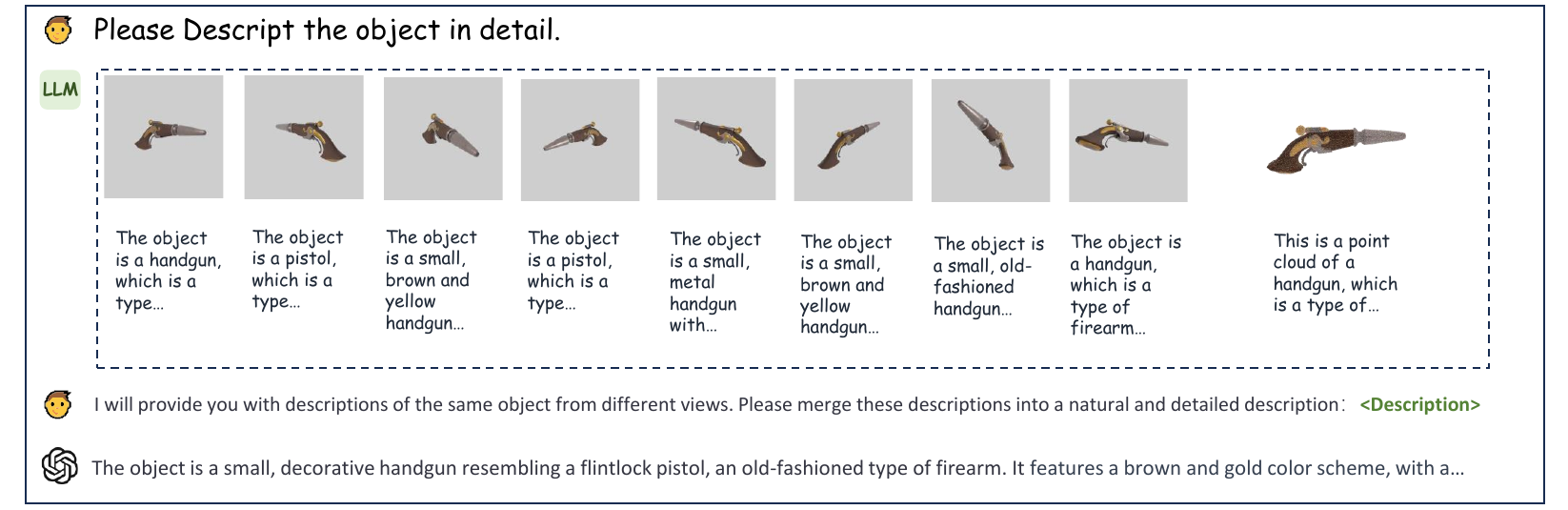}
   \caption{\textbf{The method of Cap3descript.} We input 8 perspectives render images of an object into MLLM to obtain different captions, and then use GPT for integration.}
   \label{fig:cap3descript}
\end{figure*}
\begin{figure*}[h]
  \centering
   \includegraphics[width=1\linewidth]{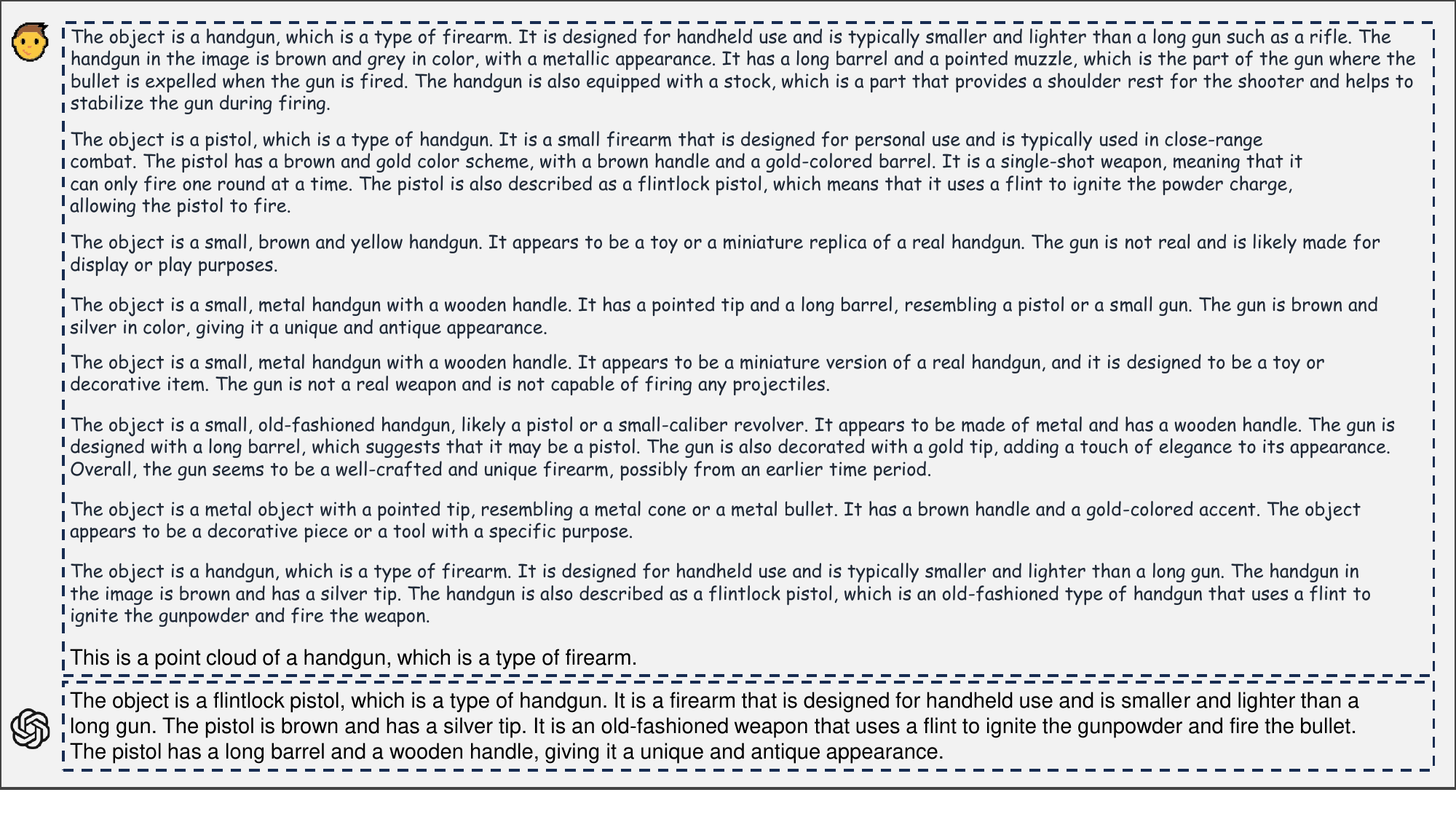}
   \caption{\textbf{A complete multi faceted description of an example.} Each angle caption is roughly same, but there may be some differences in details for specific angles.}
   \label{fig:temp}
\end{figure*}
In this section, We mainly introduce the detail for building Cap3descript.
\begin{table}
\centering
\begin{tabular}{cc}
\hline Data Source& CLIP Score\\
\hline 2D Data & 79.1\\
2D+3D Data & \textbf{79.7}\\
\hline 
\end{tabular}
  \caption{\textbf{Comparison of different data source effects.} We compared the 3D generation effect of training with normal 2D data and 2D+3D data.}
  \label{tab:2d3d}
\end{table}
The overall process is shown in \ref{fig:cap3descript}. We conducted a survey of 10 different MLLMs on the output of Cap3d render images and point cloud. We found that the output from Point-Bind\cite{guo2023point} was relatively accurate and detailed. However, due to the issue of obstruction from different angles of render images, we acquired captions of each object from eight different perspectives and the point cloud. Subsequently, we input captions from all eight angles into GPT, allowing it to merge them into a single coherent passage.\\
In our experiments, we observed that using only 3D rendered images for training the LLM-to-generator mapping block resulted in poor performance. This phenomenon could be attributed to the large amounts of blank space surrounding the 3D rendered images, as Stable Diffusion did not incorporate a significant number of such samples in its training dataset. Therefore, during training, we not only introduced 3D caption-image and description-image data but also integrated 2D text-image data for joint training. As illustrated in \cref{tab:2d3d}, it is evident that the inclusion of 3D-related data significantly enhanced the overall generation effect.
\clearpage
{
    \small
    \bibliographystyle{ieeenat_fullname}
    \bibliography{main}
}


\end{document}